%% file: main.tex
\documentclass[sigconf]{acmart}

\input{macro}

\newcommand{\tool}{\texttt{VEglue}}
\begin{document}




\title{
VEglue: Testing Visual Entailment Systems via Object-Aligned Joint Erasing
}
\author{Zhiyuan Chang}
\email{zhiyuan2019@iscas.ac.cn}
\affiliation{%
  \institution{Science and Technology on Integrated Information System Laboratory, Institute of Software Chinese Academy of Sciences; State Key Laboratory of Intelligent Game; University of Chinese Academy of Sciences}
  \city{Beijing}
  \country{China}
}

\author{Mingyang Li}
\email{mingyang2017@iscas.ac.cn}
\affiliation{%
  \institution{Science and Technology on Integrated Information System Laboratory, Institute of Software Chinese Academy of Sciences; State Key Laboratory of Intelligent Game; University of Chinese Academy of Sciences}
  \city{Beijing}
  \country{China}
}

\author{Junjie Wang}
\email{junjie@iscas.ac.cn}
\affiliation{%
  \institution{Science and Technology on Integrated Information System Laboratory, Institute of Software Chinese Academy of Sciences; State Key Laboratory of Intelligent Game; University of Chinese Academy of Sciences}
  \city{Beijing}
  \country{China}
  }

  \author{Cheng Li}
\email{licheng221@mails.ucas.ac.cn}
\affiliation{%
  \institution{Science and Technology on Integrated Information System Laboratory, Institute of Software Chinese Academy of Sciences; State Key Laboratory of Intelligent Game; University of Chinese Academy of Sciences}
  \city{Beijing}
  \country{China}
  }

\author{Qing Wang}
\email{wq@iscas.ac.cn}
\affiliation{%
  \institution{Science and Technology on Integrated Information System Laboratory, Institute of Software Chinese Academy of Sciences; State Key Laboratory of Intelligent Game; University of Chinese Academy of Sciences}
  \city{Beijing}
  \country{China}
}



\input{sec/0.abstract}
\maketitle

\input{sec/1.introduction}

\input{sec/2.background}

\input{sec/3.approach}
\input{sec/4.experiment}
\input{sec/5.result}

\input{sec/6.discussion}
\input{sec/7.related_work}
\input{sec/8.conclusion}

\bibliographystyle{ACM-Reference-Format}
\bibliography{ref}

\end{document}

%% file: macro.tex
\usepackage{soul}
\usepackage{multirow}
\usepackage{epstopdf}
\usepackage{hyperref}
\usepackage{listings}
\usepackage{fancybox}
\usepackage{graphicx}
\usepackage{subcaption}
\usepackage{paralist}
\usepackage{ragged2e}
\usepackage{enumitem}
\usepackage{tcolorbox}
\makeatletter
\usepackage{xcolor}
\usepackage{color}
\usepackage{xcolor,colortbl}
\usepackage{balance}
\usepackage{threeparttable}

\usepackage[framemethod=TikZ]{mdframed}
\usepackage{tcolorbox}
\makeatletter
\newcommand{\mybox}[1]{%
  \setbox0=\hbox{#1}%
  \setlength{\@tempdima}{\dimexpr\wd0+13pt}%
  \begin{tcolorbox}[boxrule=0.5pt, colback=white, arc=4pt,
      left=6pt,right=6pt,top=6pt,bottom=6pt,boxsep=0pt]
    #1
  \end{tcolorbox}
}
\usepackage{tikz}
\usepackage{pgfplots}
\usetikzlibrary{pgfplots.statistics,calc}
\usepackage{color}
\usepackage{array}
\usepackage{amsmath}
\usepackage{centernot}
\usepackage{xspace}
\usepackage{url}
\usepackage{verbatim}
\usepackage{wrapfig}
\usepackage{tabularx}
\usepackage{bbding}
\usepackage{pifont}
\usepackage{wasysym}
\usepackage{makecell}

\makeatletter  
\newif\if@restonecol  
\makeatother  
\usepackage[linesnumbered,ruled,vlined]{algorithm2e}
\usepackage{algpseudocode}  

%% file: sec/0.abstract.tex
\begin{abstract}

Visual entailment (VE) is a multimodal reasoning task consisting of image-sentence pairs whereby a promise is defined by an image, and a hypothesis is described by a sentence. 
The goal is to predict whether the image semantically entails the sentence. VE systems have been widely adopted in many downstream tasks.
Metamorphic testing is the commonest technique for AI algorithms, but it poses a significant challenge for VE testing. 
They either only consider perturbations on single modality which would result in ineffective tests due to the destruction of the relationship of image-text pair, or just conduct shallow perturbations on the inputs which can hardly detect the decision error made by VE systems.
Motivated by the fact that objects in the image are the fundamental element for reasoning, we propose {\tool}, an object-aligned joint erasing approach for VE systems testing.
It first aligns the object regions in the premise and object descriptions in the hypothesis to identify linked and un-linked objects.
Then, based on the alignment information, three Metamorphic Relations are designed to jointly erase the objects of the two modalities.
We evaluate {\tool} on four widely-used VE systems involving two public datasets.
Results show that {\tool} could detect 11,609 issues on average, which is 194\%-2,846\% more than the baselines.
In addition, {\tool} could reach 52.5\% Issue Finding Rate (IFR) on average, and significantly outperform the baselines by 17.1\%-38.2\%. 
Furthermore, we leverage the tests generated by {\tool} to retrain the VE systems, which largely improves model performance (50.8\% increase in accuracy) on newly generated tests without sacrificing the accuracy on the original test set.

\end{abstract}


\keywords{Visual Entailment, Metamorphic Testing, Object Perturbation}

%% file: sec/1.introduction.tex
\section{Introduction}
\label{sec:introduction}

Reasoning task is one kind of important artificial intelligence task, has made significant strides in various domains.
Precise reasoning ability is of utmost importance for advanced intelligence algorithm, as it represents a critical aspect of its intelligence level~\cite{luo2023chatgpt, Liu2023evaluate}.
Yet it is also an immensely challenging aspect.
This is particularly the case for multimodal reasoning which is due to its reliance on comprehending and integrating information from diverse modalities.



Visual Entailment (VE) is a representative task for multimodal reasoning.
It aims to reason about the relationship between an image and a sentence~\cite{ning2018visual,ning2019visual}.
Specifically, given an image as the premise, and a sentence in the text as the hypothesis, the VE task determines whether the hypothesis logically follows from the premise and outputs their relationship, i.e., entailment, neutral, or contradiction.
It affects many downstream tasks such as image captioning~\cite{peter2018bottom,xiao2022scaling}, visual question answering~\cite{stan2015vqa,Yash2019making,qing2021check}, and image-text retrieval~\cite{manh2023hada}, and is applied in many real-world scenarios such as fake news detection~\cite{Shreyash2022FACTIFY}, medically assisted diagnosis~\cite{YanakaNCK23} and robotic applications \cite{Kento2023robot}. 
The reliability of the VE systems is of great importance.
For instance, in the scenario of news content review, the automatic techniques identify fake and malicious news according to the semantic consistency relations between the containing texts and target images based on the VE systems~\cite{Suryavardan2022Factify2}.
The news with inconsistencies in texts and images, or news corresponding to some existing sensitive images is detected and warned during the content review process.
The errors in the VE systems could lead to the evasion of false facts, malicious advertisements, or hate speeches, and result in a profoundly negative societal impact.
As a vital means of quality assurance, automated testing techniques for VE systems have been less explored, and are badly desired in real-world practice.

\begin{figure}[htbp]
\centering
\includegraphics[width=8cm,height=3.0cm]{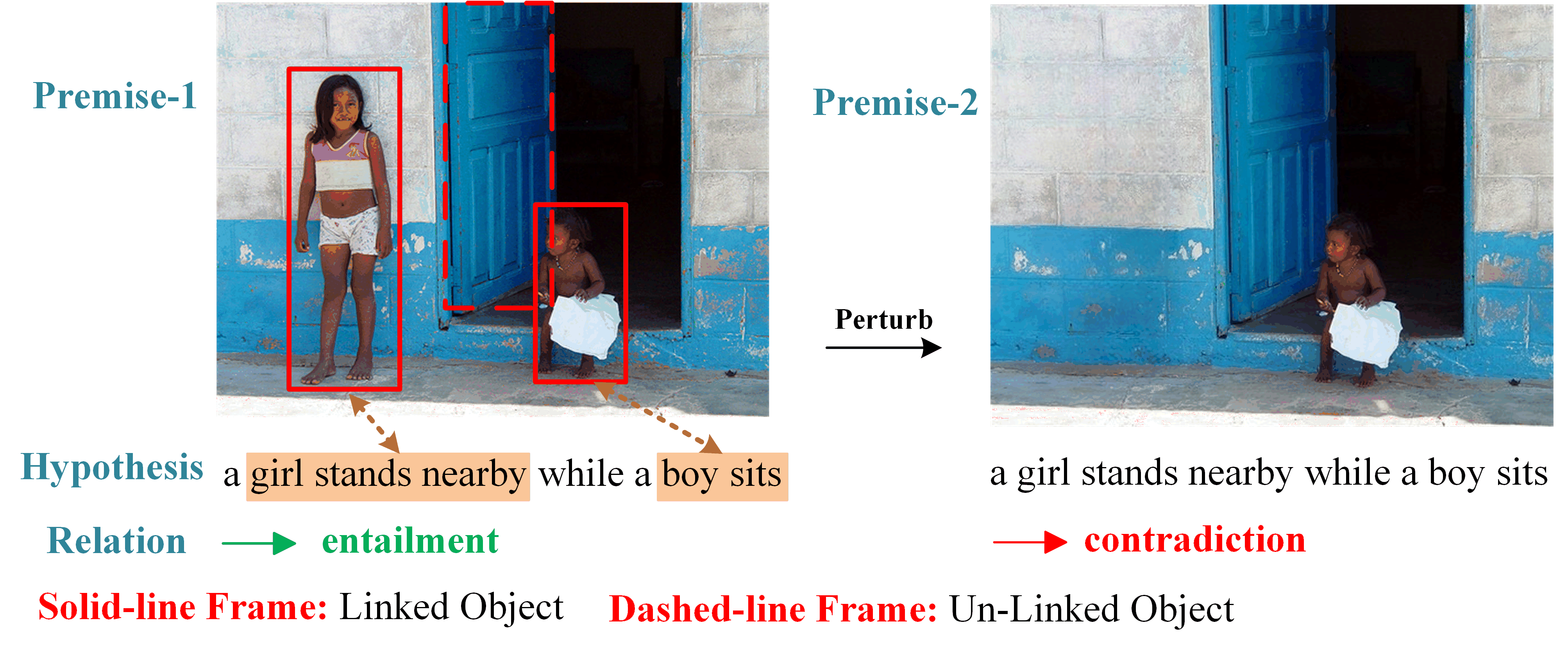}
\caption{
The origin and perturbed samples in the VE task 
}
\label{fig:motivation}
\end{figure}

When humans look at an image, they would first perceive the objects present within it. 
This holds true for VE systems as well.
As reported in existing studies~\cite{ning2019visual}, VE systems utilize objects as the fundamental building blocks for perception, reasoning, and decision.
Take Figure \ref{fig:motivation} as an example.
Given the hypothesis, the VE system would first detect the regions of objects in the premise (i.e., a girl, a boy, and a door), understand the descriptions in the hypothesis, and try to reason about whether the objects' descriptions (i.e., ``girl stands nearby'' and ``boy sits'') are consistent with the premise.
Therefore, perturbations at the object-level are likely to activate different neuronal decision-making pathways in the VE systems, and expose neurons that were previously uninvolved, uncovering potential issues.


Metamorphic testing is a framework to solve the test oracle problem due to its simplicity of concept and effectiveness in issue detection~\cite{ChenKLPTTZ18,ChenJX21,YuanWJC21}, and has been widely applied in the deep learning systems testing.
Under the framework, the core target is to design effective Metamorphic Relations (MRs) according to the characteristics of the tested systems.
Given some original samples, the new tests are generated by perturbing their inputs and obtaining the oracles using MRs.
Considering the input modalities, the general MRs for images and texts could potentially be applied in the test generation.
However, these techniques commonly fall short when dealing with the VE task due to two aspects.

\textbf{First}, the vast majority of existing studies conduct shallow perturbations to the inputs, which is to ensure that the semantics does not change substantially and keep the original output oracle unchanged. 
For example, image-oriented perturbations include blurring, snowing, zooming, etc~\cite{HendrycksD19}, and text-oriented perturbations are based on synonym substitution, back translation, simulating some character-level spelling errors, or removing some uninformative words~\cite{liu2021dialtest,0002LSBLS18}.
These perturbations are relatively subtle and shallow, which can hardly affect the decision making of the tested systems, thus resulting in low issue detection capability. 
Especially for the VE task, perturbations without changing its involved objects are not likely to challenge the system's reasoning capacity. 
\textbf{Second}, some approaches conduct deeper perturbations, such as adding objects to the images~\cite{YuZQYWH22}, inverting the semantics of text by adding negative words, or by antonym substitution~\cite{WangLGZZZYZZPWL21,JiangBTZD21}.
Yet they are oriented to a single modality.
For the VE task, the arbitrary perturbations to the input on either side may break the semantic consistency with the other side, resulting in incorrect tests.
Therefore, object-level perturbations should first consider the semantic relation between two modalities, and carefully analyze the effects on the test oracles to obtain the correct and effective MRs.

We have the following observations to motivate the design of such MRs. 
\textbf{On the one hand},
object-level perturbations should first align the object regions in the premise to the descriptions in the hypothesis.
Specifically, the premise usually contains dozens of objects, while the hypothesis tends to only mention a subset of them (denoted as \textit{linked objects}).
By making perturbations on different objects, we can derive different MRs to more thoroughly challenge the VE systems. 
For example, there are objects such as ``girl'', ``boy'', and ``door'' in the premise of Figure \ref{fig:motivation}, and the first two are linked objects while the third one is un-linked object. 
When we erase the linked object in the premise (e.g., ``girl'') and leave the hypothesis unchanged, we can derive an MR with the VE relationship changing from ``entailment'' to ``contradiction''.
Meanwhile, if we erase the un-linked object (e.g., ``door'') while keeping the hypothesis unchanged, we can derive another MR with the original VE relationship unchanged. 
This MR is also valuable because erasing the irrelevant objects can also bring noise in recognizing the linked object so as to potentially mislead the VE systems. 
\textbf{On the other hand}, we can make perturbations on single-sided (e.g., premise) or two-sided (i.e., premise and hypothesis) of the VE inputs, to derive different MRs for potentially revealing more diversified issues in the VE systems.
Take the example in Figure \ref{fig:motivation}.
When only erasing the ``girl'' from the premise and keeping the hypothesis unchanged, it can generate an MR with the VE relationship changing from ``entailment'' to ``contradiction''.
By comparison, when erasing the ``girl'' from the premise and jointly erasing the relevant description in the hypothesis (transform the original hypothesis into ``a boy sits''), we can come up with a different MR which keeps the original VE relationship unchanged. 
This type of transformation can automatically generate tests with semantic different premises and hypotheses, thus having a larger possibility in triggering different neural activate patterns and uncovering different types of issues.


Motivated by the observations, we propose an object-aligned joint erasing approach (named {\tool}) for VE systems testing.
In detail, {\tool} first identifies the objects' descriptions in the hypothesis and the objects' regions in the premise through information extraction and object detection techniques respectively.
Then, it links the objects' regions to the descriptions based on the visual grounding technique and subsequently determines the type of each object (linked or un-linked).
After that, following the framework of metamorphic testing, {\tool} designs three MRs that help generate new tests and test oracles by joint erasing on the objects and their linked descriptions, solely erasing linked objects and un-linked objects respectively.
Finally, the newly generated test inputs are fed into the VE systems and the issues are detected by comparing the predictions and corresponding test oracles.


For evaluation, we introduce two public VE datasets (SNLI-VE~\cite{ning2018visual} and e-SNLI-VE~\cite{Virginie2020esnlive}) for test generation, and four widely-used VE systems (OFA-VE~\cite{WangYMLBLMZZY22}, ALBEF-VE \cite{LiSGJXH21}, LLaVA~\cite{Liu2023llava}, GPT-4V \cite{OpenAI2023gpt4Vision}) as the tested subjects.
The performance is evaluated from two aspects, i.e., the quality of the newly generated tests and their issue detection capability.
The experimental results show that the {\tool} could detect 11,609 issues on average, which is 194\%-2,846\% more than the state-of-the-art baselines.
In addition, {\tool} could reach 52.5\% Issue Finding Rate (IFR) on average, and significantly outperform the baselines by 17.1\%-38.2\%.
Furthermore, we sample some newly generated tests as training samples to refine the two open-sourced models (OFA-VE and ALBEF-VE).
The evaluation results show the accuracy of the two models could be improved by 50.8\% without sacrificing the accuracy on the original test set.

The key contributions of this paper are as follows:
\begin{itemize}
    \item
    We propose an object-aligned joint erasing approach for testing Visual Entailment, i.e., a representative multimodal reasoning task.
    It can generate tests by understanding the object-level semantic relations between the two modalities, which can motivate future research about testing multimodal systems.


    \item 
    We evaluate {\tool} on four widely-used VE systems and two datasets, and results show that {\tool} can significantly outperform the state-of-the-art baselines with outstanding issue detection capability.
    
    \item
    By refining the open-sourced VE models with the generated tests, the accuracy on the newly generated test set could be significantly improved without sacrificing the accuracy on the original test set.
    \item
    We provide the public reproduction package\footnote{https://github.com/lsplx/VEtesting} including the tool implementation and datasets.
\end{itemize}

%% file: sec/2.background.tex
\section{Visual Entailment}
\label{sec:background}
Visual Entailment (VE) is a multimodal reasoning task that aims to understand the relationship between a given premise (serving as an image) and a hypothesis (serving as a text)~\cite{ning2018visual,ning2019visual}.
The task is framed as a ternary classification problem with the classes being entailment, contradiction, and neutral.
We will use the example in Figure \ref{fig:motivation} to elaborate on the three relationships in the following.

\begin{itemize}
    \item 
    \textbf{Entailment:} It means the content of the premise logically implies the meaning of the hypothesis. 
    The premise-1 in the example contains a girl and a boy, which is semantically consistent with what is described in the hypothesis, and therefore the relationship should be ``entailment''.
    \item 
    \textbf{Contradiction:} It signifies a direct conflict or inconsistency between a premise and a hypothesis.
    The premise-2 in the example contains only a boy, while the hypothesis describes two objects, i.e., ``a boy'' and ``a girl''.
    Therefore the relationship between the two is ``contradiction''.
    \item 
    \textbf{Neutral:} It indicates that there is no clear evidence in the premise to determine the truth of the hypothesis. 
    Using Premise-1 as an example, the corresponding hypothesis is changed to ``a girl and a boy are waiting for their father''.  
    Since the description ``waiting for their father'' in the hypothesis cannot be reflected in the premise, the relationship is considered ``neutral''.
\end{itemize}

%% file: sec/3.approach.tex
\section{Approach}
\label{sec:approach}
\begin{figure*}[htbp]
  \center{\includegraphics[width=\linewidth]{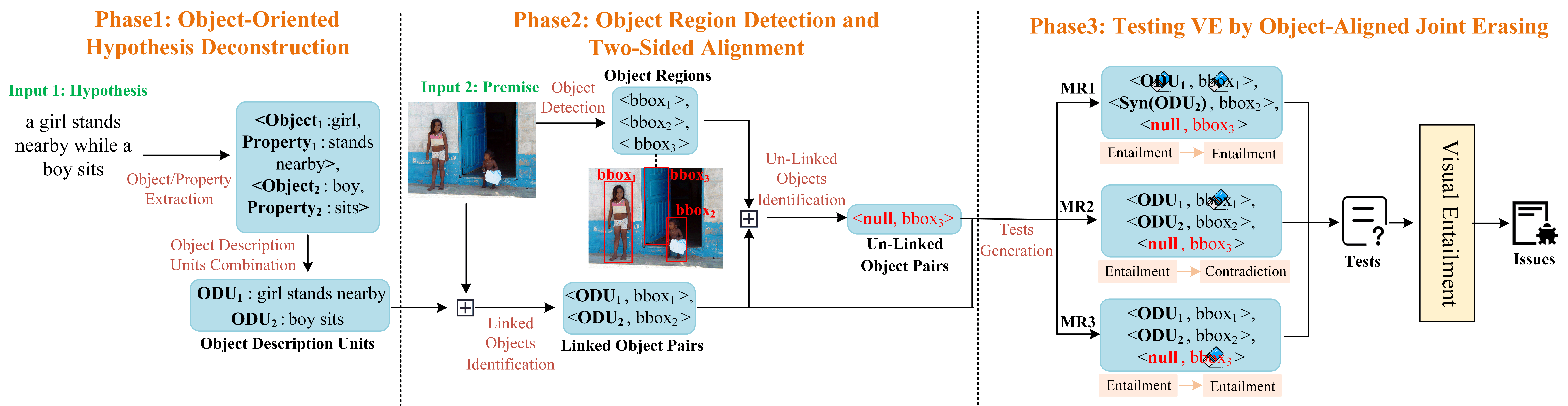}}
    \caption{
     The overview of {\tool}
     }
    \label{fig:artifacture}    
\end{figure*}

Figure \ref{fig:artifacture} shows the overview of {\tool}. 
{\tool} consists of three phases: (1) Object-Oriented Hypothesis Deconstruction, where it decomposes the hypothesis into several object descriptions; (2) Object Region Detection and Two-Sided Alignment, where it detects object regions and links objects between the premise and the hypothesis; (3) Testing VE by Object-Aligned Joint Erasing, where it designs three MRs by object-aligned joint erasing for testing VE system.
In the following subsections, we will first describe three MRs (Section \ref{sec:approach_MR}). 
Next, we provide the details of Phase 1 (Section \ref{sec:obj_decomposition}), Phase 2 (Section \ref{sec:approach_Combination}), and Phase 3 (Section \ref{sec:approach_Melting}).

 \subsection{Object-Aligned Metamorphic Relations}
\label{sec:approach_MR}
{\tool} proposes three MRs to help test generation. 
We introduce their concepts below.

\textbf{MR1: Perturbation on the object and its linked description.} 
Erase an object in the premise and its linked object descriptions in the hypothesis, and apply the synonym transformation to the remaining object descriptions.
The corresponding relationship label remains ``entailment''\footnote{For ``neutral'' and ``contradiction'' relationships, the object descriptions are inaccurately linked with the objects, thus this MR is not suitable for them.}.

\textbf{MR2: Perturbation on linked object.} 
Erasing an object in the premise while keeping its linked object descriptions in the hypothesis would change the original ``entailment'' relationship into ``contradiction''.


\textbf{MR3: Perturbation on un-linked object.} 
Erasing an un-linked object in the premise while keeping the hypothesis would remain the ``entailment'' relationship unchanged.

\subsection{Object-Oriented Hypothesis Deconstruction}
\label{sec:obj_decomposition}
A hypothesis may involve multiple objects (e.g., ``girl'' and ``boy'' in Figure \ref{fig:motivation}), and each object can usually be described with associated property (e.g., ``stands nearby'' and ``sit'' in the hypothesis). 
We denote an ``\textbf{object description unit}'' as a specific object together with its property (or properties) (e.g., ``girl stands nearby'', ``boy sits'').
Accurately deconstruct the hypothesis into the object description units is the basis for the follow-up linking with the premise. 
Nevertheless, these description units may be connected by auxiliary words (e.g., conjunctions ``while'') or may have overlapping words with each other (e.g., ``man flying a kite'' and ``boy flying a kite'' in the hypothesis ``There is a man and a boy flying a kite''), which exerts challenges in the decomposition.
To tackle it, we formulate this task as extracting and reassembling the fine-grained elements (i.e., object and property), with details shown below.


\subsubsection{\textbf{Object/property Extraction}}
\label{subsec_entity_extraction}
This step aims to extract object/property from the hypothesis, which serve as the elements for deriving the object description units.
Table 1 shows the definitions and examples of the object and property. 

\input{tab/entity_category.tex}

{\tool} employs ChatGPT\footnote{https://openai.com/blog/chatgpt}, which is a commonly-used Large Language Model (LLM), for object/property extraction from the hypothesis.
ChatGPT is a generative model that interacts with users in the form of question and answer, and has shown considerable text reasoning capabilities~\cite{Zhong2023ChatGPT} and achieves impressive performance on the information extraction task~\cite{Bo2023entity}.
To employ it for the object/property extraction, it's crucial to craft suitable questions, i.e., prompts.
To this end, {\tool} builds on the prompt used in the previous study~\cite{Bo2023entity}, and enhances it by adding 10 extra input-output samples to the prompt aiming at helping ChatGPT achieve better extraction performance in our context.
The template for the prompt is shown below.

\begin{tcolorbox}
\textbf{Prompt:} 
Please extract the object/property in the following sentence and output it in JSON format.
Here are examples: 

\textbf{Example1:} 

\textit{Input:} the man is aiming his rifle at something. 

\textit{Output:}[\{ "object": "man" , "property": "is aiming his rifle at something"\}, \{ "object": "rifle" , "property": ""\} ] 

...

\textbf{Example10:} ...
\newline
The following is the input for object/property extraction: 

\textit{Input:} 
{[Input hypothesis];} 
  \textit{Output:}
\end{tcolorbox}

For input-output samples, {\tool} selects the samples with longer text length from the candidate sets (the candidate sets are described in Section \ref{sec:experiment_design}).
The underlying idea is that the longer the length of the hypothesis, the more object description units it contains and the more complex the sentence structure, which might guide ChatGPT to learn complex examples so that it can better handle simple ones.
Given a hypothesis for object/property extraction, it is added to the placeholder ``[Input hypothesis]'' in the prompt template before being fed into ChatGPT. 
Then, ChatGPT outputs object/property in a format that aligns with our sample output.

\subsubsection{\textbf{Object Description Units Combination}}
This step aims to combine the object/property extracted in the previous step to generate coherent and complete object description units.
After extracting the object/property from the ChatGPT, {\tool} then gets the index of the objects and properties in the hypothesis and arranges object/property such that the one with the smaller index is placed first, and vice versa.
Finally, {\tool} obtains the object description units in the input hypothesis, the example is shown in Phase 1 of Figure \ref{fig:artifacture}.

\subsection{Object Region Detection and Two-Sided Alignment}
\label{sec:approach_Combination}
This phase aims at aligning the objects in the premise and hypothesis, i.e., linking each \textit{object description unit} with the object region in the premise.  
{\tool} designs two steps: (1) Object Detection, where it detects all objects and marks the object regions in the premise; (2) Object Linking, where {\tool} identifies the linked objects among them based on the description units.
The following introduces the details of two steps.

\subsubsection{\textbf{Object Detection}}
We mentioned in Section \ref{sec:introduction}, only when we can accurately separate the linked objects and un-linked objects, we can make up diversified MRs to more thoroughly test the VE system. 
In order to determine the linked and un-linked objects in the premise, we first find all candidate ones from the premise. 
Hence, this step aims to detect all objects in the premise and determine the coordinate information of each object. 

To achieve it, given the input premise,  {\tool} employs the object detection technique \cite{ZhaoZXW19} to locate the objects in the premise.
The goal of object detection is to localize and classify each object, and in this paper, we only use the localization information output by the model.
We use the state-of-the-art object detection algorithm~\cite{0097LL000NS23} to recognize the object regions in the input premise, and the final output of the model is the coordinate information of object regions in the premise.
The coordinate information is represented by a bounding box (``\textbf{bbox}''), which contains four values, the horizontal and vertical coordinates of the top-left corner of the region and the horizontal and vertical coordinates of the bottom-right corner.
Finally, {\tool} obtains bbox information for all objects detected in the premise.
As shown in the object regions in Phase 2 of Figure \ref{fig:artifacture}, three bbox information are detected from the input premise.

\subsubsection{\textbf{Object Linking}}
\label{sec:link}
This step aims to find the linked objects and un-linked objects. 
To identify linked objects, we use a Visual Grounding (VG) model to identify the object regions for all the object description units in the hypothesis. 
Visual Grounding~\cite{MaoHTCY016} is to localize the object regions on the image according to a textual expression.
The inputs to the VG model are a premise and a hypothesis, and the output is the bbox information linking to the hypothesis, i.e., the coordinate position of the region in the premise.
As shown in Phase 2 in Figure \ref{fig:artifacture}, this phase outputs Linked Object pairs, where $bbox_{1}$ links to the ``girl stands nearby'' object description unit and $bbox_{2}$ links to ``boy sits'' object description unit.
To identify un-linked objects, {\tool} compares the object regions detected by object detection with the linked object determined by VG.
Specifically, for each object region detected by the object detection model, {\tool} compares the degree of overlap between it and the linked object region.
The extent of overlap between the two regions can be quantified by calculating the Intersection over Union (IoU) value~\cite{EveringhamGWWZ10}.
IoU is a metric that quantifies the similarity between two regions by comparing the area of their intersection with the area of their union.
The IoU metric is calculated using the following formula:
\begin{equation}
IoU(A, B) = \frac{Area(A \cap B)}{Area(A \cup B)}
\end{equation}
where $A$ represents the predicted region, $B$ represents the ground truth region, $A \cap B$ is the intersection of the two regions, and $A \cup B$ is their union.
The IoU value ranges from 0 to 1, where 0 indicates no overlap and 1 represents a perfect match between the predicted and ground truth regions.
If the IoU value is lower than the pre-defined threshold (set as 0.1 based on empirical experience), it means that the object region is un-linked object.
After identifying all object regions, {\tool} obtains Un-Linked Object pairs in the premise, as illustrated by the door corresponding to $bbox_{3}$ in Figure \ref{fig:artifacture}.

\subsection{Testing VE by Object-Aligned Joint Erasing}
\label{sec:approach_Melting}

Based on the fine-grained relations established in previous sections, this phase aims to implement MRs (described in Section \ref{sec:approach_MR}) to generate tests for testing the VE system by object-aligned joint erasing.

\subsubsection{\textbf{Implementation of MR1}}
For MR1, {\tool} erases both the linked object regions and object description units. 

For erasing the object description units from the hypothesis, the challenge is how to handle the overlapping situation between two object description units. 
We design the following strategies to tackle it. 
For each object description unit, if its property value contains the object value of the remaining object description units, then only the object value of this object description unit is taken to combine with the other object description units.
Take a hypothesis (``the man is aiming his rifle at something'') as an example.
There are two object description units in the hypothesis. 
The first object description unit has the object ``man'' with the property ``is aiming his rifle at something'', and the second has ``rifle'' as the object with an empty property.
If the second object description unit of ``rifle'' is erased, then ``rifle'' should not appear in the first object description unit.
Consequently, we only consider the object ``man'' from the first object description unit for subsequent combination, excluding its property value ``is aiming his rifle at something''.
Then, it introduces perturbations via semantically-equivalent synonym substitutions for the combined object description units to generate a new hypothesis.

For erasing the object region in the premise, the challenge is how to generate a correct and natural image. 
To tackle this, we first check whether the target object can be erased without influencing the image's naturalness based on the overlapping status, and for the satisfied ones, we conduct the erasion with the image inpainting technique. 
In detail, 
{\tool} first assess whether the region of the erased object overlaps with the remaining objects.
If the region of overlap is significant, it may impact the surrounding image regions when erasing the image object, consequently affecting the correctness and naturalness of the generated tests.
The degree of overlap is evaluated by the IoU value.
If the IoU value exceeds a predetermined threshold (set as 0.1 as previous section), the object could not be erased from the image.
For the object that can be erased, {\tool} first masks the object region and then employs image inpainting techniques to repair the masked region.
The goal of image inpainting is to naturally fill in object regions within an image, which requires the inpainting model to understand the structural and content information of images.
{\tool} uses a state-of-the-art image inpainting model~\cite{SuvorovLMRASKGP22}. 
The inputs to the model are the original image and a binary mask image, in which the object region is white and the rest of the region is black.
The model erases the object in the premise and fills the missing region with background contexts.
The Premise-2 of Figure \ref{fig:motivation} shows the new premise after erasing the girl. 
Since the premise and the hypothesis erase the linked objects, the relationship between them remains ``entailment''.


\subsubsection{\textbf{Implementation of MR2}}
For MR2, {\tool} erases only the object regions linked with the object description units.
{\tool} follows the same procedure as MR1 about erasing the object region in the premise to do that. 
As shown in Figure \ref{fig:motivation}, Premise-2 is a new premise generated by MR2, and its hypothesis keeps unchanged.
There is an obvious contradiction between the content in the premise and the semantics of ``girl stands nearby'' in the hypothesis, thus the relationship between the two is ``contradiction''.

\subsubsection{\textbf{Implementation of MR3}}
For MR3,{\tool} erases only the object region un-linked to the object description unit.
Specifically, for each region that is not linked with an object description unit, {\tool} erases it from the premise using masking and image inpainting techniques following the procedure of MR1.
For example, if the door in Premise-1 of Figure \ref{fig:motivation} is erased and the hypothesis remains unchanged, the content in the premise can still imply the meaning of the hypothesis, and the relationship between the two is still ``entailment''.

After test generation through MRs, {\tool} inputs them into the VE system for testing. 
For each test, if the relationship predicted by the VG system violates the MR, the test is treated as an ``\textbf{issue}''.
Take an example of a test generated by MR2, given the perturbed premise and its paired hypothesis, the VE system predicts the relationship between the two.
MR2 is the transformation of the original ``entailment'' relationship into ``contradiction''.
If the prediction fails to yield ``contradiction'', then MR2 is violated, and thus the test is considered as an issue being detected in the VE system.

%% file: tab/entity_category.tex
\begin{table}[t] \huge
\caption{The definitions and examples of object and property}
\label{tab:entity_categories}
\resizebox{\columnwidth}{!}{
\begin{tabular}{ccc}
\toprule
\textbf{Category}    &   \textbf{Definition} &   \textbf{Example}  \\   
\midrule
\multirow{2}{*}{{Object}}  &  the main subject described in   &  ``girl'' in the  \\
& the object description & hypothesis of Figure \ref{fig:motivation} \\ \midrule
\multirow{2}{*}{{Property}}   &  the modified expression   &   ``stands nearby'' in  \\
& for the object & the hypothesis of Figure \ref{fig:motivation}  \\

\bottomrule
\end{tabular}
}
\end{table}

%% file: sec/4.experiment.tex
\section{Experiment}
\label{sec:experiment}
\subsection{Research Questions}

\textbf{RQ1: How effective is {\tool} in detecting issues for VE systems?}
Taking four widely-used VE systems as the tested subjects, we investigate the issue detection effectiveness and advantage of the proposed approach.



\textbf{RQ2: Can the three MRs contribute to issue detection capability of {\tool}?}
We evaluate the effectiveness of each MR in {\tool} in terms of issue detection separately.

\textbf{RQ3: Can the tests generated by {\tool} help improve the performance of the VE models?}
We investigate whether the performance of the open-sourced VE models could be improved with the tests generated by {\tool}.

\subsection{Dataset}
We employ two public datasets (i.e. SNLI-VE and e-SNLI-VE) which are been commonly-used as the benchmarks for the VE task.
Specifically, SNLI-VE contains a training set, a validation set, and a test set with 529,527, 17,858, and 17,901 samples (i.e., premise-hypothesis pairs and corresponding relationship labels) respectively, where the training set and validation set are used to optimize the model hyper-parameters during the training phase, and the test set is used to assess its performance after the model is finalized.
For e-SNLI-VE, it contains a training set with 529,526 samples, an evaluation set with 17,555 samples, and a test set with 17,572 samples.
In our study, we leverage the test sets of the two dataset as the inputs to {\tool} for test generation.

\subsection{Tested VE Subjects}
Our evaluation is based on four VE systems, including two representative VE models (i.e., OFA-VE~\cite{WangYMLBLMZZY22}, ALBEF-VE~\cite{LiSGJXH21}) and two large visual language models (i.e. LLaVA~\cite{Liu2023llava}, GPT-4V~\cite{OpenAI2023gpt4Vision}).
Specifically, OFA-VE and ALBEF-VE are fine-tuned from the two widely-used pre-trained multi-modality models (i.e. OFA and ALBEF respectively), and have demonstrated impressive performance in the VE benchmark datasets including SNLI-VE and e-SNLI-VE.
Besides the two fine-tuned models, we also introduce two large visual language models (VLMs), i.e. LLaVA and GPT-4V.
They are popularly-used in the real world to solve many multimodal tasks such as VE and Visual Question Answering (VQA).
Specifically, LLaVA is the open-sourced VLM, and we use the LLaVA-1.5 version which obtains the state-of-the-art performance across 11 benchmarks including VE~\cite{Liul2023report}.
GPT-4V is the latest commercial VLM released by OpenAI, and has shown remarkable capabilities on a wide of multimodal tasks~\cite{Yang2023Dawn}.
In our study, we also employ the newly generated tests to detect their issues when dealing with the VE task.

\subsection{Experiment Design}
\label{sec:experiment_design}
To answer RQ1, for object/property extraction step, we use ChatGPT to extract object/property from the input hypothesis.
Specifically, we utilized the March 1st version of GPT-3.5 (gpt-3.5-turbo-0301).
For the object detection step, we use the DINO~\cite{0097LL000NS23}, a popular object detection model, to detect object regions in the input premise.
It achieves 63.3 AP on the COCO 2017 dataset, which demonstrates its precise object detection capabilities and strong scalability.
Then, for the object linking step, we feed the object description units and the input premise into the Visual Grounding model.
We use the OFA-VG~\cite{WangYMLBLMZZY22} as the VG model, which demonstrates a highly satisfactory performance, achieving an impressive accuracy of 94.0\% on the RefCOCO dataset.
For implementing MRs, we use the Lama model~\cite{SuvorovLMRASKGP22} to erase the object regions in the premise, which achieves an impressive Fréchet Inception Distance score of 0.63 on the Places dataset and is wildly-used in the testing approach \cite{YuZLYHH23}.
Finally, for each dataset, {\tool} generates the tests using the samples in the test set, and the generated tests are sent to the VE system for issue detection.
For evaluation, first, we evaluate the valid rate of the generated tests by manual checking.
Of all the generated tests, we selected 100 tests by random sampling and a team with one senior researcher and two Ph.D. students manually checked whether the sampled tests were valid.
For each test (consisting of a premise-hypothesis pair and its test oracle), it is considered ``\textbf{valid}'' if all three members in the team generally believe that the test oracle correctly describes the relationship between premise-hypothesis pair.  
Second, we evaluate the issue detection performance of the generated tests.
Specifically, we count the number of tests generated by the approach, the number of tests that detected issues, and the proportion of these issue-detecting tests relative to the total generated.
Moreover, we introduce two state-of-the-art text testing approaches and a state-of-the-art image perturbation approach (illustrated in Section \ref{sec:exe_baselines}), as the baselines.
To answer RQ2, for each dataset, we first obtain linked and un-linked object pairs after using object/property extraction and combination, object detection, and object linking.
Then the tests are generated using MRs in the {\tool}, respectively, resulting in three distinct sets of tests, which correspond to MR1, MR2, and MR3.
These sets of tests are then fed into the VE system for issue detection.
Same as RQ1, we also evaluate the quality and issue detection capability of generated tests under each MR.
The final result is the average of the performance on the two datasets.



To answer RQ3, we use the retraining strategy on two research models (i.e. OFA-VE and ALBEF-VE) to explore whether the newly retrained VE model can improve the performance on tests generated by {\tool}.
As for LLaVA and GPT-4v, they are multimodal large models. 
Considering that their training process requires massive computational resources and even GPT-4v is not open-sourced, we did not evaluate them in this RQ.

Following the experimental setting in the previous study~\cite{Gupta20,HeMS20,ChenJX21} for enhancing the multimodal models, we divide the generated tests into a performance-improving set and an evaluation set in the ratio of 8:2 by random sampling.
After that, we add the performance-improving set to the original training set to retrain the VE models and evaluate the performance on the evaluation set.
In addition, we evaluate the performance of the retrained model on the original test set to determine whether the retraining strategy has compromised the model's reasoning capacities on the original test set.


\subsection{Baselines}
\label{sec:exe_baselines}
We do not find any testing techniques specific to the VE task.
Considering the correlation between premises and hypotheses in the VE task, the following approaches have been adopted as baselines to fully preserve their correlation.

\textbf{TextFlint~\cite{WangLGZZZYZZPWL21}.}
It is the state-of-the-art testing approach for the Textual Entailment (TE) task, in which both premise and hypothesis are text.
In TextFlint, tests are crafted through an MR that substitutes words in the hypothesis with their antonyms, thereby altering the oracle to ``contradiction''.
In our study, we reuse the package provided by the paper\footnote{https://github.com/textflint/textflint}.

\textbf{CAT~\cite{0004Z0HP022}.}
It is the state-of-the-art testing approach for the translation system under the framework of metamorphic testing.
In CAT, the tests are generated using the MR that perturbs the inputs by word substitution and keeps the associated oracle unchanged.
In our study, its public package\footnote{https://github.com/zysszy/CAT.} is leveraged for test generation.

Besides the above baselines are oriented to the texts, we additionally employ a state-of-the-art image perturbation as a baseline.

\textbf{Feature-based Image Perturbation (FIP)~\cite{HendrycksD19}. } 
It is proposed for image recognition related tasks.
FIP perturbs the pixel values of an image by incorporating noise, blur, weather, and digital effects. 
For each type of effect, FIP has four perturbation methods, and we selected one as a representative to serve as the baseline for our analysis.
The core idea is to introduce noise to the original image from various aspects, preserving its semantic content.
In our study, we use the public package of FIP\footnote{https://github.com/bethgelab/imagecorruptions} to perturb the input premises and keep the original relationship unchanged as test oracles.

\subsection{Evaluation Metrics}

To measure the quality of the generated tests, we introduce  \textit{Valid tests Rate (VTR)}, the ratios of the valid generated tests to all the generated tests (Details of validity evaluated are described in Section \ref{sec:experiment_design}).
To measure the issue detection capability of the generated tests, we use three widely-used metrics.
1) \textit{Generated Number (GNUM)}, which counts how many tests are generated by the testing approach.
2) \textit{Issue Number (INUM)}~\cite{shen2022natural}, which counts how many issues are detected by generated tests with the VE system.
3) \textit{Issue Finding Rate (IFR)~\cite{YuanWJC21,LiuFY022,wang2023mttm}}, which is the ratio of the number of tests that detect issues to the total number of generated tests.
Additionally, to measure the performance of the VE system before and after enhancement in RQ3, we use Accuracy, the ratio of correct predictions to all predictions, which is the commonly-used metric to evaluate the performance in the VE task~\cite{ning2019visual,cao2022alignve}.

%% file: sec/5.result.tex
\section{Results}
\label{sec:result}
\subsection{Answering RQ1}
\input{tab/effectiveness_RQ1}
Table \ref{tab:RQ1_performance} shows the detected Issue Number (INUM) and Issue Finding Rate (IFR) of {\tool} and baselines on four VE systems, as well as VTR and GNUM of the generated tests.
In general, on the OFA-VE, ALBEF-VE, and LLaVA systems, {\tool} detects 9,695,  11,293, and 13841 issues respectively.
Moreover, {\tool} achieves IFR of 47.4\%, 55.2\%, and 69.4\% on three VE systems, and significantly outperforms the state-of-the-art baselines by 14.7\%-44.3\%. 

Compared to the TextFlint, {\tool} detects 2,846\% more issues and achieves 16.8\% higher in IFR on average, which shows the advances of {\tool} in both detected issue number and detection efficiency.
Due to the simplicity of the perturbation (only involves antonyms substitution), the number of generated tests is quite limited. 
Accordingly, the quantity of detected issues is also smaller.
Furthermore, it ignores the effects of objects and may not expose system issues more efficiently.
For the quality of the generated tests, TextFlint shows better performance than {\tool}.
Since it only does antonym substitution for certain words (i.e., gender-related words), the probability of generating erroneous tests is relatively low. 
{\tool} achieves 92.6\% VTR, which is comparably close to that of TextFlint.
Taking into account its issue detection capabilities, {\tool} is able to expose more issues.
Based on these issues, after filtering erroneous tests by manual review or automated content review techniques, which is more practical for quality evaluation and subsequent improvement of the VG model in practice.



Compared to the CAT, {\tool} detects 194\% more issues and achieves 35.0\% higher in IFR on average.
CAT generates tests by word substitution, the IFR is also relatively low (22.3\% on average).
However, compared to TextFlint, CAT is capable of substituting universally typed words and therefore can generate far more tests and thus get higher INUM than TextFlint.
For the quality of the generated tests, {\tool} is 12.3\% higher than CAT on VTR.
This is because CAT employs BERT model for word substitution. 
Due to the absence of corresponding synonym constraints during the substitution process, BERT sometimes replaces the current word with an antonym, causing a semantic reversal in the entire sentence and generating erroneous tests.


Compared with FIP, {\tool} detects 228\%-308\% more issues, and the IFR value is 37.4\%-41.3\% higher. 
FIP achieves perturbation by adding noise to the image globally.
 It often faces a dilemma: if the perturbation is too subtle, it fails to challenge the model (e.g., Snow); if it's too aggressive, it may render the image's object unrecognizable (e.g., Zoom), leading to the generation of invalid tests.
 In contrast, {\tool} can perceive objects within images and ensures semantic consistency between image perturbations and oracles based on MRs, thereby demonstrating superior performance.

We evaluated our approach using OpenAI's newest release, GPT-4V. 
Remarkably, {\tool} achieves an IFR of 37.8\%, and outperforms the baseline by 15.7\%-28.2\%, which indicates the superior capabilities of {\tool} in issue detection for the relatively mature commercial systems.
In addition, compared with the other three VE systems, the lowest IFR on GPT-4V indicates its superior image-text comprehension and higher robustness.
The results also reveal that the text-based perturbation baselines (TextFlint and CAT) exhibit a more pronounced issue detection capability than the image-based perturbation baseline (FIP). 
This suggests that as a form of discrete data, text perturbations are more likely to induce fluctuations in the performance of large visual language models.

\subsection{Answering RQ2}
\input{tab/ablation_RQ2}

Table \ref{tab:rq2} presents the Issue Number (INUM) and the Issue Finding Rate (IFR) for each MR in {\tool} on four VE systems, as well as
VTR and GNUM of the generated tests.
We first observe the issue detection capability of each MR on OFA-VE, ALBEF-VE, and LLaVA.
It can be found that the Issue Number of MR1 is the lowest compared to the other two MRs.
This is mainly because MR1 is to simultaneously erase the linked object in the premise and associated object description in the hypothesis.
The hypothesis suitable for MR1 typically requires two or more object description units, to ensure that at least one piece of object description is retained in the hypothesis.
The nature of MR1 makes it generate fewer tests compared to the other MRs.

MR2 is to erase linked objects in premises, and it exhibits the highest INUM and IFR values among all MRs.
Compared to MR1, MR2 only performs object perturbation in premises. 
The number of the generated tests is not restricted by the number of object description units in the hypothesis, thus it exhibits better generality and could generate more tests.
The main difference between MR2 and MR1 is that MR2 leads to a change in the relationship, while MR1 maintains the original relationship.
The results show that the IFR value of MR2 is 25.6\% higher than that of MR1, and the IFR of the baseline SwapAnt and MR-NR approaches, which also change the relationship in the baselines, are higher than those of the rest of the baselines.
This indicates that the perturbations capable of altering the relationship are more effective in detecting issues.

MR3 is to erase un-linked objects in premises.
Since a premise often contains far more objects than object description units in a hypothesis, there are lots of un-linked object regions that can be erased. 
Therefore, MR3 could typically generate multiple tests from a single original test and achieve the highest GNUM among all MRs.
 Furthermore, erasing un-linked objects indirectly affects the recognition of linked objects, it could potentially mislead the VE system.
The results show that it detected 5,322 issues on average, outperforming all baselines.

As for GPT-4V, the IFR of MR1 is much higher than that of MR2 and MR3. 
Considering that MR1 contains both image and text perturbations, while MR2 and MR3 are image-only perturbations, the results indicate that perturbations in image-text union can be more effective in detecting the issues of GPT-4V, which is already robust to image perturbations.

For the quality of the generated tests, the results show that the VTR of MR1 falls short compared to the other two MRs, with MR3 exhibiting the highest VTR among the three MRs.
This is mainly because MR1 perturbs both the premise and the hypothesis, and MR2 and MR3 only perturb the premise.
The perturbation of the hypothesis introduces a certain amount of cumulative errors.
The erroneous tests generated by three MRs are mainly due to the fact that the introduced VG model does not correctly link the objects in the premise and hypothesis, which results in a cumulative error.
However, the employed VG model can reach an accuracy of 94.0\%, which is sufficient to ensure reliability in most scenarios.
Future enhancements to the VG model can further elevate the ratio of valid tests generated by {\tool}.



\subsection{Answering RQ3}
In this RQ, we investigate whether the tests generated by {\tool} can aid in improving the accuracy of the VE model.
Table \ref{tab:rq3} shows the accuracy for the original/retrained VE models under tests generated by {\tool} on two datasets (The experimental settings described in Section \ref{sec:experiment_design}).
The results show that on the two VE models, the average accuracy of the retrained model is improved by 47.2\% and 54.4\% over the original model, respectively.
It indicates that the retrained model can effectively fix the issues existing in the original model to improve accuracy.
Besides, the performance of the retrained model on the original test set on the two datasets is not much different from the original model (the accuracy difference is within 1.0\%), which means that the retraining does not affect the original performance of the model.

\input{tab/retrain_RQ3}


%% file: tab/effectiveness_RQ1.tex
\begin{table}[t]

  \caption{The quality and the issue detection ability of the tests generated by {\tool} and baselines}
  \label{tab:RQ1_performance}
  
\resizebox{0.5\textwidth }{!}{
\begin{threeparttable}
\begin{tabular}{cc|ccccccccc}
\toprule
\multirow{2}{*}{\textbf{VE Systems}} &
  \multirow{2}{*}{\textbf{Metric}} & 
  \multirow{2}{*}{\textbf{{\tool}}} &
  \multirow{2}{*}{\textbf{TextFlint}} &

  \multirow{2}{*}{\textbf{CAT}} &
  \multicolumn{4}{c}{\textbf{FIP}}\\     \cline{6-9} 

 &
   &
    &
    &
 
     &

  \multicolumn{1}{c}{\textit{\textbf{Impulse}}} &
  \textit{\textbf{Zoom}} &
  \multicolumn{1}{c}{\textit{\textbf{Snow}}} &
    \multicolumn{1}{c}{\textit{\textbf{Contrast}}} 
  \\ \midrule
\multirow{2}{*}{{OFA-VE}} &
  \textit{INUM} &

  \multicolumn{1}{c}{\textbf{9695}} & 
  \multicolumn{1}{c}{318} & 

  \multicolumn{1}{c}{3085} &

2110 &

  2838&
   
  \multicolumn{1}{c}{2155} &

  \multicolumn{1}{c}{1951} \\ 
  
&\textit{IFR} &

  \multicolumn{1}{c}{\textbf{47.4\%}} & 
  \multicolumn{1}{c}{32.7\%} & 

  \multicolumn{1}{c}{17.4\%} &

  11.9\% & 
  
  16.0\% &
   
  \multicolumn{1}{c}{12.2\%} &
   
  \multicolumn{1}{c}{11.0\%} \\

   \midrule
\multirow{2}{*}{{ALBEF-VE}} &
  \textit{INUM} &
\multicolumn{1}{c}{\textbf{11293}} & 
  \multicolumn{1}{c}{389} & 

  \multicolumn{1}{c}{3191} &

   2305  &
  3015 &
   
  \multicolumn{1}{c}{2455} &
   
  \multicolumn{1}{c}{2128} \\ 

    & \textit{IFR} &
\multicolumn{1}{c}{\textbf{55.2\%}} & 
  \multicolumn{1}{c}{40.0\%} & 

  \multicolumn{1}{c}{18.0\%} & 
  
   13.0\% &

  17.0\% &
   
  \multicolumn{1}{c}{13.9\%} &
   
  \multicolumn{1}{c}{12.0\%} \\ 

  \midrule
\multirow{2}{*}{{LLaVA}} &
  \textit{INUM} &
\multicolumn{1}{c}{\textbf{13841}} & 
  \multicolumn{1}{c}{474} & 

  \multicolumn{1}{c}{5585} &

   4592  &
  4771 &
   
  \multicolumn{1}{c}{4504} &
   
  \multicolumn{1}{c}{4451} \\ 

    & \textit{IFR} &
\multicolumn{1}{c}{\textbf{69.4\%}} & 
  \multicolumn{1}{c}{48.7\%} & 

  \multicolumn{1}{c}{31.5\%} & 
  
   25.9\% &

  26.9\% &
   
  \multicolumn{1}{c}{25.5\%} &
   
  \multicolumn{1}{c}{25.1\%} \\ 
  \midrule
\multirow{2}{*}{{GPT-4V\tnote{*}}} &
  \textit{INUM} &
\multicolumn{1}{c}{\textbf{340}} & 
  \multicolumn{1}{c}{66} & 

  \multicolumn{1}{c}{64} &

   29  &
  46 &
   
  \multicolumn{1}{c}{33} &
   
  \multicolumn{1}{c}{36} \\ 

    & \textit{IFR} &
\multicolumn{1}{c}{\textbf{37.8\%}} & 
  \multicolumn{1}{c}{22.1\%} & 

  \multicolumn{1}{c}{21.5\%} & 
  
   9.6\% &

  15.4\% &
   
  \multicolumn{1}{c}{11.0\%} &
   
  \multicolumn{1}{c}{12.0\%} \\
  \midrule

  \multirow{2}{*}{{-}} &\textit{VTR} &

 \multicolumn{1}{c}{92.6\%} & 
  \multicolumn{1}{c}{\textbf{98.3\%}} & 

  \multicolumn{1}{c}{80.3\%} &

  91.3\% & 
  
  62.7\% &
   
  \multicolumn{1}{c}{89.3\%} &
   
  \multicolumn{1}{c}{76.4\%} \\
   &\textit{GNUM} &

 \multicolumn{1}{c}{\textbf{20453}} & 
  \multicolumn{1}{c}{972} & 

  \multicolumn{1}{c}{17730} &

  17730 & 
  
  17730 &
   
  \multicolumn{1}{c}{17730} &
   
  \multicolumn{1}{c}{17730} \\

  \bottomrule
\end{tabular}
\begin{tablenotes}
      \item[*]
      The API opened on November 6, 2024, and is limited to 100 calls per day. Given the requirements of testing approaches, this number is significantly less than our needs. 
      For the tests generated by each testing approach, we randomly selected 100 tests from it to verify its issue detection capability.
      The experiment was repeated three times. 
      IFR used the average value as the final result, and INUM used the sum of the three experiments as the final result.
    \end{tablenotes}
  \end{threeparttable}
}
\end{table}

%% file: tab/ablation_RQ2.tex
\begin{table}[t] 
\tiny
\caption{The quality and the issue detection ability of tests generated by each MR}
\resizebox{0.9\columnwidth}{!}{

\label{tab:rq2}

\begin{tabular}{cc|ccc}
\toprule
\textbf{VE Systems} &\textbf{Metrics} & \textbf{MR1} & \textbf{MR2} & \textbf{MR3} \\ \midrule
\multirow{2}{*}{{OFA-VE}} &{INUM} &   674     &   4830   &  4192 \\ 

 &{IFR}  & 43.9\% &   75.8\% & 22.5\%   \\

 \midrule

\multirow{2}{*}{{ALBEF-VE}} &{INUM}  &   707    & 6059 &   4527    \\ 

 &{IFR}  &   46.1\%     & 95.1\%  &   24.3\%    \\ 
 
 \midrule
 \multirow{2}{*}{{LLaVA}} &{INUM}  &   1332    & 5263 &   7247    \\ 

 &{IFR}  &   86.8\%     & 82.6\%  &   38.9\%    \\ 
 \midrule
 \multirow{2}{*}{{GPT-4V}} &{INUM}  &   192    & 106 &   43    \\ 

 &{IFR}  &   64.0\%     & 35.4\%  &   14.2\%    \\ 
 \midrule
 
 \multirow{2}{*}{{-}} &{VTR}  & 89.5\% &   91.0\% & 97.2\%   \\
   &{GNUM}  & 1535 &   6371 & 18631   \\

\bottomrule
\end{tabular}

  }
 
\end{table}

%% file: tab/retrain_RQ3.tex
\begin{table}[t]
\caption{The accuracy of original/retrained models on two datasets}
\label{tab:rq3}
\resizebox{\columnwidth}{!}{
\begin{tabular}{l|ccc}
\toprule
\textbf{} & \textbf{SNLI-VE} & \textbf{e-SNLI-VE} & \textbf{Average} \\ \midrule

Original OFA-VE &   51.1\%     &   54.1\%     &   52.6\%     \\

Retrained OFA-VE & 99.7\% (48.6\%$\uparrow$)  & 99.9\% (45.8\%$\uparrow$)  &   99.8\% (47.2\%$\uparrow$)    \\

\midrule

Original ALBEF-VE &   43.5\%     &   46.2\%     &   44.9\%     \\

Retrained ALBEF-VE  & 99.4\% (55.9\%$\uparrow$)  & 99.0\% (52.8\%$\uparrow$) &   99.2\% (54.4\%$\uparrow$) \\


\bottomrule
\end{tabular}
}
\end{table}

%% file: sec/6.discussion.tex
\section{Discussion}
\label{sec:discussion}

\begin{figure*}[t]
  \center{\includegraphics[width=0.9\linewidth,height= 7cm]{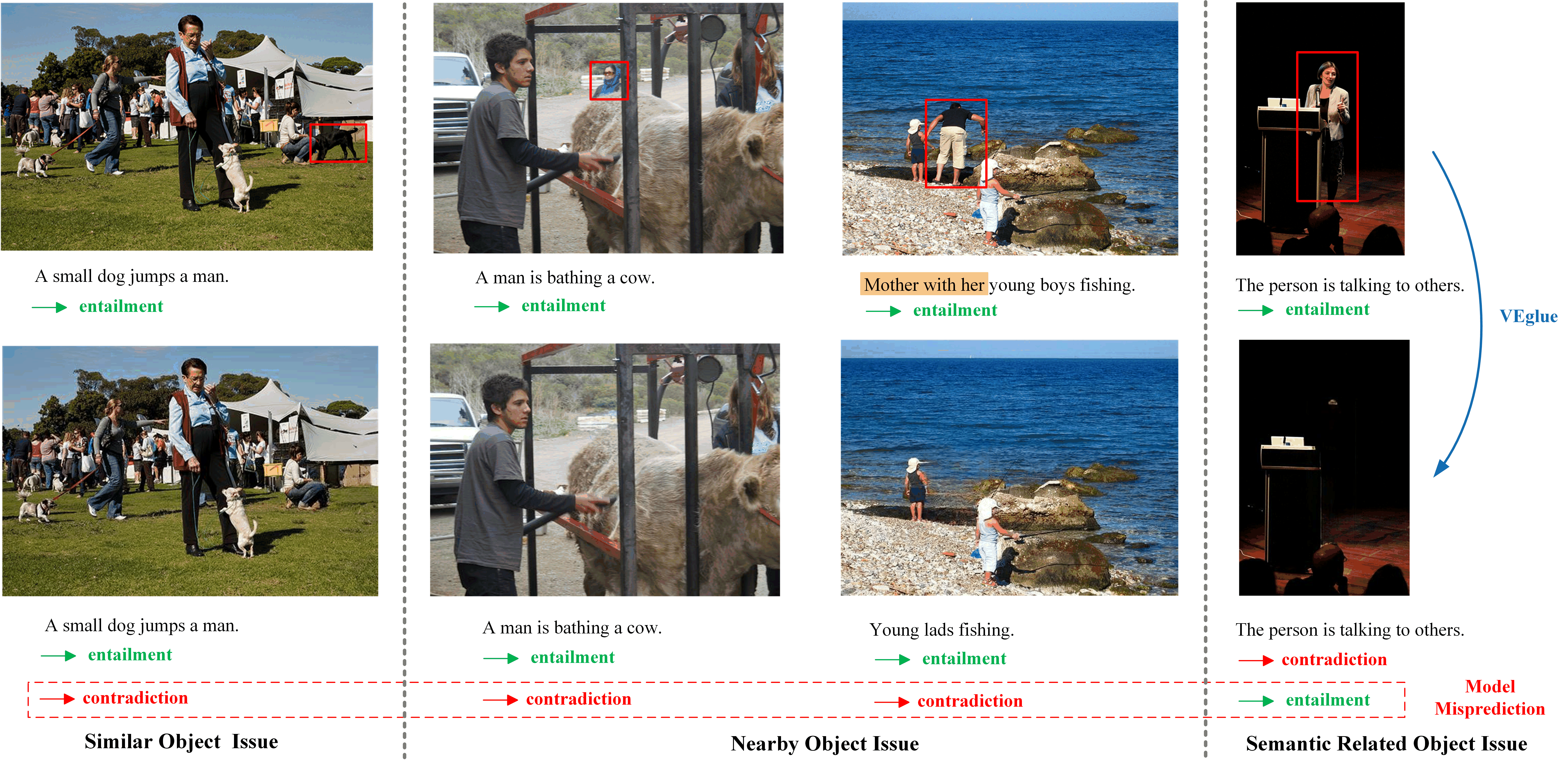}}
    \caption{
     Three types of detected issues in VE system testing
    }
    \label{fig:badcase}    
\end{figure*}

\subsection{Unveiling the Mystery of the Detected Issues}


We explore three types of issues while testing to give a systematic and intuitive understanding of issues detected by {\tool}.
In our study, we found that the exposed issue in the four tested subjects share significant similarities, so we will take one of them as a case for illustration.
 Figure \ref{fig:badcase} illustrates four representative issue types for OFA-VE.
The four tests (premise-hypothesis pair and its oracle relationship) in the top row of Figure \ref{fig:badcase} come from the original test set and four new tests generated by {\tool} are shown in the bottom row. 
The dashed box in the figure represents the relationships incorrectly predicted by OFA-VE (i.e., Model Misprediction).

The first type of issue involves a Similar Object misleading the VE system. Specifically, removing an un-linked object that bears a high resemblance but is unrelated to the linked object could lead the VE system to make an erroneous decision.
In the first column of Figure \ref{fig:badcase}, {\tool} erases the region of the black dog, that bears a strong resemblance to ``a small dog'', from the original premise.

The second issue type is Nearby Object misleading the VE system.
Specifically, erasing an object in the premise that is spatially adjacent to the linked object would result in the VE system making incorrect decisions.
This type of issue can be detected by MR1 (for erasing linked objects) and MR3 (for erasing un-linked objects).
Regarding MR3, taking the example in the second column of Figure \ref{fig:badcase}, the woman (an un-linked object) closely after the cow (a linked object) in the premise is erased, which may influence the feature embedding of the linked object in the VE system and makes the incorrect prediction.
As for MR1, as illustrated in the third column of Figure \ref{fig:badcase}, the mother (a linked object) that is close to another lined object in the premise (``young boys fishing'') is erased.
In contrast to MR3, the linked object description unit is also removed, and synonym substitution is applied to the remaining object description units (changing ``boys'' to ``lads'').
In light of the issues identified above, we believe that enhancing the VE system's ability to distinguish between closely related objects in the premise is one way to improve the system's reasoning capabilities.

The third issue type is Semantic Related Object misleading the VE system,i.e., the remaining object still has a semantic relation with the perturbed hypothesis which would mislead the VE system.
As shown in the fourth column of Figure \ref{fig:badcase}, the talking person in the premise is erased, and there is still a semantic relation between the remaining object (i.e., the off-stage listener) and the object description unit (``person talking to other''). 
This would mislead the system to reason incorrectly.



\subsection{Performance of Object/Property Extraction}

The quality of the tests generated by the {\tool} is influenced by the performance of object/property extraction in Section \ref{subsec_entity_extraction}.
We design an LLM based technique to automatically conduct the extraction. 
Here we present its performance to indicate the reliability of our approach. 

We use three widely-used metrics to evaluate the performance, i.e., Precision, Recall, and F1~\cite{EbertsU20,LampleBSKD16}.  
Of all hypotheses after extraction, we randomly sampled 100 samples, and a member of the review team (same as the team described in Section \ref{sec:experiment_design}) conducted the manual check.
The object/property extraction results are shown in Table \ref{tab:entity_extraction}, which presents the extraction performance for each category with and without in-context learning.
On the one side, the results show a promising overall extraction performance in 95.6\% F1, which implies the credibility of the designed model. 
On the other side, compared with extraction without in-context learning, LLM with in-context learning could increase 8.4\% F1.

\input{tab/entity_extraction}

\subsection{Sensitivity to Perturbed Regions in Premise}
\input{tab/local_whole_disscusion}
Previous image-oriented perturbation techniques perturbed the entire image globally, which are unable to perceive key elements in the premise. 
While {\tool} can accurately identify objects in the premise. 
Building upon this, we further explore the sensitivity of previous perturbation techniques to the perturbed regions in the premise.
Table \ref{tab:dis_local_global} shows the Issue Finding Rate for different perturbations occurring on object and global region for four VE systems, where impulse, zoom, snow, and contrast correspond to the perturbations in FIP.
Additionally, we introduce object earsing (the column ``Earsing'' in Table \ref{tab:dis_local_global})  for comparison in investigating the advantages of the core perturbation in {\tool}. 

The results show that shallow image perturbations at the object regions could reach 17.2\% IFR on average, which is 1.1\% higher than the IFR value on the global region.
This suggests that shallow perturbations in the object region are also a potential approach for detecting issues in the VE system.
In addition, the erasing perturbation is 35.3\%, 45.5\%, 33.6\%, and 11.1\% higher than the IFR of shallow perturbations on the object region for the four VE systems, respectively.
This indicates that erasing perturbation of the object region is more effective in detecting VE system issues.
However, shallow perturbations and erasing are two different granularities of perturbations, leading to different categories of issues being detected, and future work that combines the two perturbations is a potential approach to detecting a wider range of issues.


\subsection{Potential Multimodal Scenarios for {\tool}}
\label{dis:general}

{\tool} is a testing approach designed in the scenario of VE, which has achieved superior performance. 
 Future work could extend its application to a broader range of multimodal scenarios from the following perspectives.

First, {\tool} can be applied to scenarios that test VQA systems.
The input to the VQA scenario is an image and a descriptive query asking about the content of the image, and the output is a textual response to the query~\cite{MalinowskiF14}.
In testing VQA systems, MR3 in VEglue can be used directly, and MR1 and MR2 can be applied to some of the scenarios in the VQA task.
Specifically, for MR3 in VEglue, it first extracts object descriptions from the query. VEglue then identifies and erases objects in the image that are not mentioned in the query. With the modified image and original query, the VQA system's responses should still be consistent.
Additionally, MR1 removes the object and its associated description to assess the VQA system's handling of queries with multiple object descriptions. Take the query ``Are there apples and bananas in the image?'' as an example. Erasing ``bananas'' from the query and the corresponding image region should not alter the VQA system's answer. 
Conversely, MR2 erases only the object region; hence, if both apples and bananas are present, the VQA system's response should switch from ``yes'' to ``no''.
Second, {\tool} can also be used for testing VLMs.
Nowadays, the research of VLMs receives more and more attention~\cite{Liu2023llava,bai2023qwen}, and testing VLMs is of significant importance.
The inputs and outputs of VLMs are consistent with VQA, so {\tool} can be used to test them as well.

\subsection{Threats to Validity}
\textbf{{External Validity}}.
The external threats are related to the generalization of the proposed approach.
First, we experiment with two publicly available datasets, which can alleviate this issue.
Second, we experiment with four widely-used VE systems, which can alleviate this issue.
Third, 
with uncomplicated adaptation (details in Section \ref{dis:general}), {\tool} can be applied to the testing of other multimodal systems, which can alleviate this issue.

\textbf{{Internal Validity}}.
The internal threats relate to experimental errors and biases.
First, the Visual Grounding model can only localize a region in a premise, which could lead to mislocalization if the object in the object description unit is plural (e.g., two girls in the image). 
We will first determine whether the object number in the object description unit is plural or not.
 If it is, we avoid inputting it into the Visual Grounding model, thereby mitigating this threat.
Second, the image inpainting technique used by the approach may affect the rest of the region when repairing the object region.
Before object erasing, we will first determine whether the region largely overlaps with the rest of the target region.
If the degree of overlap exceeds a specified threshold, the object will not be erased, which can help to mitigate this threat.

%% file: tab/entity_extraction.tex
\begin{table}[t]\Huge
\caption{The performance of object/property extraction}
\label{tab:entity_extraction}
\centering
\resizebox{\columnwidth}{!}{
\begin{tabular}{cc|cccc}
\toprule
\textbf{Category}  & \textbf{Mode}  & \textbf{Precision} & \textbf{Recall} & \textbf{F1} \\ \midrule
\multirow{2}{*}{Object} & with in-context learning & 96.0\%  & 96.0\%  & 96.0\%     \\
  & without in-context learning & 90.7\%  & 90.7\%  & 90.7\%     \\ \midrule
\multirow{2}{*}{Property}&  with in-context learning& 94.4\%  & 95.8\%  & 95.2\%    \\

&  without in-context learning& 81.0\%  & 86.7\%  & 83.7\%    \\

\bottomrule
\end{tabular}
}
\end{table}

%% file: tab/local_whole_disscusion.tex
\begin{table}[t]
\caption{The Issue Finding Rate (IFR) for image perturbations on object/global region}
\label{tab:dis_local_global}
\centering
\resizebox{\columnwidth}{!}{
\begin{tabular}{cc|ccccc}
\toprule
\textbf{VE Systems}  & \textbf{Region}  & \textbf{Impulse} & \textbf{Zoom} & \textbf{Snow} & \textbf{Contrast} & \textbf{Earsing}    \\ \midrule
\multirow{2}{*}{OFA-VE} & Object & 12.5\%  & 17.5\%  & 12.3\% & 13.0\% & 49.1\%      \\
  & Global & 11.9\%  & 16.0\% & 12.2\% & 11.0\% & -     \\ \midrule
\multirow{2}{*}{ALBEF-VE}&  Object& 12.9\%  & 17.2\%  & 14.8\% & 12.0\%  & 59.7\%   \\

&  Global & 13.0\%  & 17.0\%   & 13.9\% & 12.0\% & -  \\
\midrule
\multirow{2}{*}{LLaVA}&  Object& 25.9\%  & 31.2\%  & 25.8\% & 25.9\% & 60.7\%    \\

&  Global   & 25.4\%   & 26.9\% & 25.5\%  & 25.1\% & - \\
\midrule
\multirow{2}{*}{GPT-4V}&  Object& 10.8\%  & 16.3\%  & 13.5\% & 14.3\% & 24.8\%    \\

&  Global   & 9.6\%   & 15.4\% & 11.0\%  & 12.0\% & - \\

\bottomrule
\end{tabular}
}
\end{table}

%% file: sec/7.related_work.tex
\section{Related Work}
\label{sec:related work}




\subsection{NLP Systems Testing}
As NLP technology is rapidly developing, numerous testing approaches have been proposed to test NLP systems. 
For example, Liang et al. \cite{0002LSBLS18} design three perturbation strategies for the DNN-based text classifiers, namely insertion, modification, and removal.
 Liu et al. \cite{LiuF021} employs transofrmation-specific (e.g. Back Translation, Word Insertion, and Synonym Replacement) perturbations between the transformed sentences and seed data to detect issues in the dialogue system.
The above methods, however, only provide a superficial level of perturbation and do not induce perturbations at the object level within the hypothesis, which would limit the issue detection capability in the VE task.
Shen et al. \cite{shen2022natural} propose a sentence-level mutation based metamorphic testing technique for Question answer software.
They detect issues in Question answer software by adding irrelevant statements to the question or context.
As the premises and hypotheses are linked, any added sentences should be considered in the context of the premises. 
Therefore this approach cannot be used directly for the VE task.

\subsection{CV Systems Testing}
In the literature, several testing approaches have been proposed to test CV systems.
 \cite{HendrycksD19} assess image classifier robustness on common corruptions and perturbations by creating corrupted images using various techniques, including noise, blur, weather-related effects, and digital manipulations. 
These perturbations are relatively subtle and shallow, which can hardly affect the issue detection capability.
Geirhos et al. \cite{GeirhosRMBWB19} propose the method of background style migration to perturb the image for testing the object recognition systems.
Specifically, they use neural networks to separate and recombine content and style from different images, creating a new image that combines the content of one image with the artistic style of another.
However, under the VE task, the change in color of the objects in the premises due to style migration would affect the correspondence between the original premises and hypotheses.
This makes it impossible to determine changes in the test oracle, and thus this approach is not applicable to the VE task.

\subsection{Multimodal Systems Testing}
As multimodal tasks are increasingly used in real-life applications, many testing approaches have been proposed to test multimodal systems.
Yu et al. \cite{YuZQYWH22} propose a metamorphic testing approach to validate Image Caption systems.
They create an object corpus by extracting objects from images and insert them into new images using resizing and repositioning algorithms. Yet, they overlook the contextual fit of the inserted objects with the background, leading to many unrealistic images unsuitable for the VE task.
Yuan et al. \cite{YuanWJC21} design MRs to test Visual Question Answering systems from the perspective of low-level perception.
Specifically, they jointly transform the image and question into one or a set of sub-images and sub-questions.
Then, they check whether the answer to the original image and question satisfies MRs with the composed answers of transformed images and questions.
The approach does not need to consider the relation with text when dividing images into sub-images.
Therefore, it is not appropriate for the VE task where premises and hypotheses are linked.

%% file: sec/8.conclusion.tex
\section{Conclusion}
\label{sec:conclusion}
We propose {\tool}, an object-aligned joint erasing approach for Visual Entailment systems testing.
Drawing inspiration from the fact that objects in the image are the fundamental element for reasoning, the core idea behind MR design is to focus on the object-level perturbations within the premises and hypotheses.
{\tool} first identifies the object descriptions in the hypothesis and the object regions in the premise, and links the object regions to the descriptions.
After, following the framework of metamorphic testing, {\tool} designs three MRs that generate new tests and test oracles by joint erasing on the objects and their linked descriptions, solely erasing linked objects and un-linked objects respectively.
The evaluation shows {\tool} could outperform the state-of-the-art testing approaches for issue detection on four VE systems.
In addition, by retraining the VE models with the tests generated by the {\tool}, the performance of the VE models on newly generated tests can be significantly improved.
In the future, we will design more MRs on the objects for the VE task and investigate whether the core idea of {\tool} could be adapted to test various multimodal tasks.